\documentclass[sigconf]{acmart}

\usepackage{layouts}
\usepackage{array}
\usepackage{enumitem}

\AtBeginDocument{%
  \providecommand\BibTeX{{%
    \normalfont B\kern-0.5em{\scshape i\kern-0.25em b}\kern-0.8em\TeX}}}

\setcopyright{none}
\copyrightyear{2026}
\acmYear{2026}

\renewcommand\footnotetextcopyrightpermission[1]{}

\usepackage{multirow}
\usepackage{pgfplots}
\pgfplotsset{compat=1.18}
\usepackage{graphicx}
\usepackage{subfig}
\usepackage[font=small,skip=0pt]{caption}
\usepackage{algorithm} 
\usepackage{algpseudocode} 
\usepackage{tabularx,booktabs,caption}
\usepackage{arydshln}
\usepackage{stfloats}
\usepackage{pgfplots}
\usepackage{tikz}
\usepgfplotslibrary{groupplots}
\pgfplotsset{compat=1.18}

\begin{document}
\title[FedeRICo]{FedeRICo: Federated Region-Influenced Coupling for Traffic Flow Prediction}
%%Federated Bi-Gradient Alignment for Traffic Flow Prediction with Cross-Client Communication
%%
%% The "author" command and associated commands define the authors
%% and their affiliations.

\author{Fermin Orozco}
\affiliation{%
  \institution{University of Exeter}
  \city{Exeter}
  \country{United Kingdom}}
\email{fo260@exeter.ac.uk}

\author{Man Luo}
\affiliation{%
  \institution{University of Exeter}
  \city{Exeter}
  \country{United Kingdom}}
\email{M.Luo@exeter.ac.uk}

\author{Johan Wahlström}
\affiliation{%
  \institution{University of Exeter}
  \city{Exeter}
  \country{United Kingdom}}
\email{J.Wahlstrom@exeter.ac.uk}

%%
%% Short author list for page headers.
\renewcommand{\shortauthors}{Orozco et al.}

\begin{abstract}
Urban traffic forecasting often relies on information distributed across stakeholders who may be unable to share raw data due to privacy or commercial constraints, motivating federated spatial-temporal approaches. In such federated settings, each client observes traffic over a distinct sensor subgraph with its own spatial topology and temporal dynamics, leading to significant heterogeneity across clients. Existing federated spatial-temporal methods typically rely on model parameter aggregation and provide limited mechanisms for recovering spatial dependencies across client boundaries. This introduces two key limitations. Specifically, parameter aggregation across heterogeneous graph domains tends to dilute client-specific representations, while road network partitioning breaks the propagation of traffic dynamics across client boundaries. To address these challenges, we propose FedeRICo, a federated traffic forecasting framework that combines gradient-level collaboration with boundary-aware residual communication. FedeRICo employs a dual-branch forecasting architecture in which a globally guided branch captures transferable forecasting structure, while a private residual branch preserves client-specific corrections and incorporates boundary residual signals. The global branch is coordinated through gradient alignment across all clients, enabling collaborative optimisation without destructive parameter interference. To recover cross-client spatial dependencies, boundary messages are extracted through a trend-residual decomposition that suppresses periodic structure and communicates only transient spatial-temporal residual signals between physically adjacent clients. Experiments across four real-world traffic forecasting benchmarks demonstrate that FedeRICo consistently outperforms state-of-the-art federated spatial-temporal baselines while maintaining competitive training runtime.
\end{abstract}




%\received{*}
% \received[revised]{*}
% \received[accepted]{*}

\maketitle

\section{Introduction}

Spatial-temporal forecasting is an essential part of modern Intelligent Transport Systems; the ubiquitous deployment of infrastructure and sensors have led to large-scale data that can be leveraged to optimise urban services \cite{service_ref_1_crosscityfew, service_ref_2_citytraffic, service_ref_3_MoE} and enhance safety \cite{safety_ref_1_accident, safety_ref_2_eval, safety_ref_3_CARPG}. To model the inherent interdependencies within this data, deep spatial-temporal models implement a combination of GNNs \cite{GNN}, and RNN \cite{RNN} or attention \cite{attention} based architectures to model the spatial and temporal relationships, respectively \cite{STGCN, DCRNN, GWN, AGCRN, STDN}. Yet these approaches assume a centralised setting, where all the data is centrally available on a single system. In real-world settings, there may be multiple organisations, each with their own independent dataset, that may not be able to centrally aggregate their data due to privacy or commercial limitations. This motivates decentralised data-driven methods for optimising distributed traffic forecasting models. 

Federated Learning (FL) \cite{FedAvg} enables training under this constraint since it does not require centralisation of data. Organisations may collaborate in a decentralised manner to train optimal models without needing to share their raw sensitive data. Several Federated methods have been proposed for traffic flow prediction \cite{FedGRU, CNFGNN, FedGTP, pFedCTP, FedDis}. Such approaches collaborate primarily through parameter aggregation on shared model components \cite{pFedCTP, FedDis, CNFGNN, FedGRU}, while others learn cross-client interactions through latent representations \cite{FedGTP}.

The spatial and temporal heterogeneity across client datasets poses difficulties for FL. First, naive parameter aggregation across clients can dilute client-specific representations built during local training. Decoupling spatial and temporal parameters as an alternative, risks losing the integrated spatial-temporal modelling that gives these architectures their predictive power. Second, partitioning a connected road network across clients disrupts spatial dependencies at partition boundaries, limiting the cross-client information flow that boundary nodes may otherwise contribute.

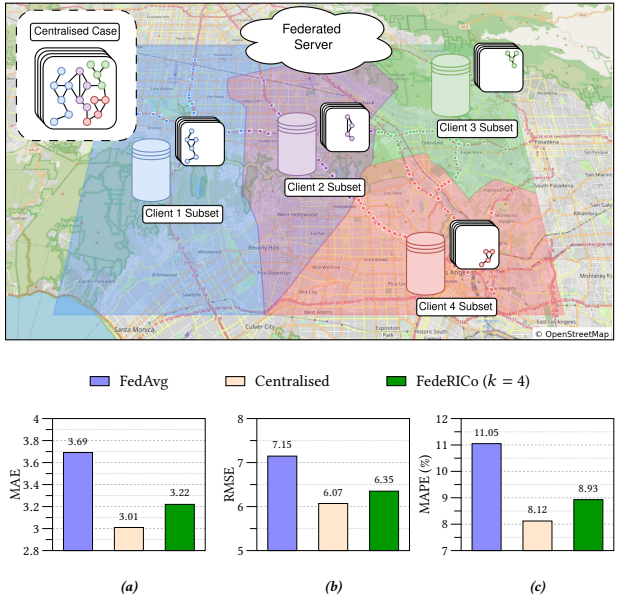
\begin{figure}[t]
\centering
\scriptsize

% ==========================================
% TOP: The Visual Map
% ==========================================
\includegraphics[width=0.95\columnwidth]{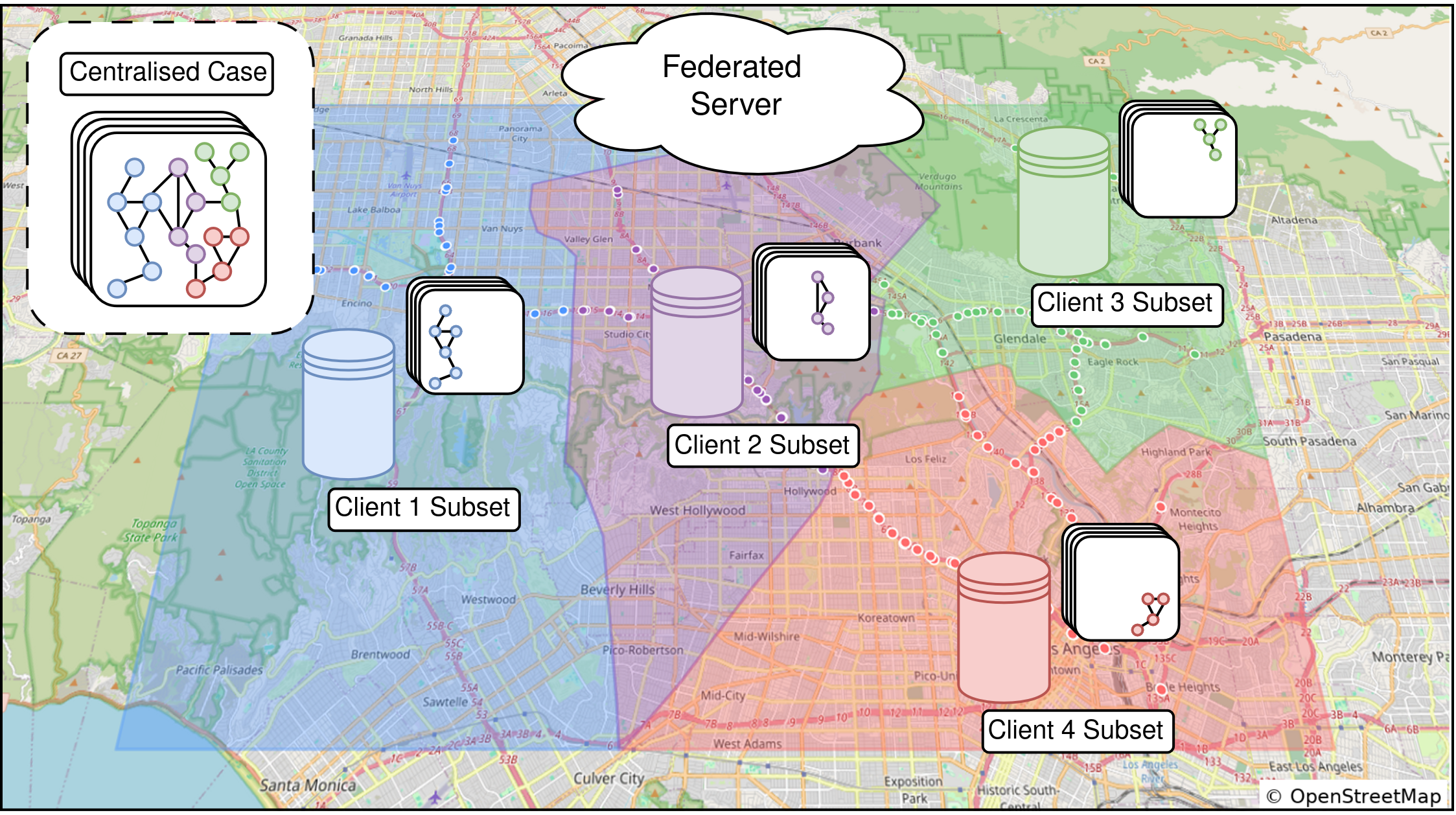}

\vspace{0.35cm}

% =========================
% Centered manual legend
% =========================
\makebox[0.95\columnwidth][c]{%
\begin{tabular}{@{}c@{\hspace{0.75cm}}c@{\hspace{0.75cm}}c@{}}
\tikz{\node[draw=black, fill=blue!45, minimum width=0.25cm, minimum height=0.12cm, inner sep=0pt] {};}\hspace{0.08cm} FedAvg
&
\tikz{\node[draw=black, fill=orange!20, minimum width=0.25cm, minimum height=0.12cm, inner sep=0pt] {};}\hspace{0.08cm} Centralised
&
\tikz{\node[draw=black, fill=green!60!black, minimum width=0.25cm, minimum height=0.12cm, inner sep=0pt] {};}\hspace{0.08cm} FedeRICo ($k=4$)
\end{tabular}%
}

\vspace{0.20cm}

% =========================
% BOTTOM: Three-panel plot
% =========================
\makebox[0.95\columnwidth][c]{%
\resizebox{0.95\columnwidth}{!}{%
\begin{tikzpicture}

\begin{groupplot}[
    group style={
        group size=3 by 1,
        horizontal sep=0.75cm
    },
    width=3.8cm,
    height=3.5cm,
    ybar,
    /pgf/bar width=12pt,
    /pgf/bar shift=0pt,
    xmin=0.5,
    xmax=3.5,
    enlarge x limits=false,
    clip=false,
    xtick=\empty,
    axis lines=box,
    axis line style={line width=0.25pt},
    grid=both,
    minor tick num=1,
    major grid style={gray!55, dash pattern=on 0.8pt off 0.8pt, line width=0.16pt},
    minor grid style={gray!35, dash pattern=on 0.6pt off 0.6pt, line width=0.12pt},
    tick align=outside,
    tick pos=left,
    tick style={black, line width=0.1pt},
    yticklabel style={font=\fontsize{5.0}{4.0}\selectfont, xshift=1pt},
    ylabel style={font=\fontsize{6.0}{5.0}\selectfont, yshift=-0.1cm},
    label style={font=\fontsize{6.0}{5.0}\selectfont},
    title style={
        at={(0.5,-0.30)},
        anchor=north,
        font=\fontsize{6.0}{5.0}\selectfont\bfseries\itshape
    },
    nodes near coords,
    nodes near coords style={
        font=\fontsize{4.8}{4.8}\selectfont,
        anchor=south,
        yshift=1pt
    },
    every node near coord/.append style={
        /pgf/number format/fixed,
        /pgf/number format/precision=2
    }
]

% =========================
% (a) MAE
% =========================
\nextgroupplot[
    ylabel={MAE},
    ymin=2.80, ymax=4.00,
    ytick={2.80, 3.00, 3.20, 3.40, 3.60, 3.80, 4.00},
    title={(a)}
]
\addplot[fill=blue!45, draw=black] coordinates {(1,3.6912)};
\addplot[fill=orange!20, draw=black] coordinates {(2,3.01)};
\addplot[fill=green!60!black, draw=black] coordinates {(3,3.22)};

% =========================
% (b) RMSE
% =========================
\nextgroupplot[
    ylabel={RMSE},
    ymin=5.00, ymax=8.00,
    ytick={5.00, 6.00, 7.00, 8.00},
    title={(b)}
]
\addplot[fill=blue!45, draw=black] coordinates {(1,7.1468)};
\addplot[fill=orange!20, draw=black] coordinates {(2,6.07)};
\addplot[fill=green!60!black, draw=black] coordinates {(3,6.35)};

% =========================
% (c) MAPE (%)
% =========================
\nextgroupplot[
    ylabel={MAPE (\%)},
    ymin=7.00, ymax=12.00,
    ytick={7.00, 8.00, 9.00, 10.00, 11.00, 12.00},
    title={(c)}
]
\addplot[fill=blue!45, draw=black] coordinates {(1,11.05)};
\addplot[fill=orange!20, draw=black] coordinates {(2,8.12)};
\addplot[fill=green!60!black, draw=black] coordinates {(3,8.93)};

\end{groupplot}
\end{tikzpicture}%
}%
}

\vspace{0.10cm}

\caption{Performance and spatial partitioning summary on METR-LA. \textbf{Top:} Spatial partitioning into $k=4$ localized clusters. \textbf{Bottom (a--c):} Evaluation of our proposed framework against federated and centralised baselines, demonstrating superior predictive accuracy against the federated setting, approaching the centralised model performance.}
\label{fig:Intro_Figure}

\end{figure}

Figure \ref{fig:Intro_Figure} illustrates the challenge of FL under real-world urban traffic constraints wherein organisations or municipalities may have ownership of spatially distributed infrastructure, and hence would have a partitioned view of the full network. Furthermore, they will possess different spatial distributions, and as a result, will learn spatial-temporal models over different underlying graph structures.

The proposed Federated Region-Influenced Coupling (FedeRICo) framework addresses both of these difficulties through two complementary mechanisms. First, gradient-based training collaboration is implemented, replacing parameter averaging with a learning-based alignment that preserves client specific representations. Second, a boundary messaging protocol exchanges transient, event-informative signals between physically adjacent clients to recover some of the cross-client information flow at partition boundaries. These mechanisms are realised through a dual-branch architecture, wherein a global branch captures shared dynamics through gradient alignment among all clients, while a local branch models client-specific patterns and the residual error. Figure \ref{fig:Intro_Figure}(b) shows that this approach substantially outperforms naive parameter averaging (FedAvg) and approaches the centralised STDN model \cite{STDN} upper bound, in which all data is idealistically aggregated to a single system. These results indicate that gradient-based alignment is a more effective collaboration mechanism than parameter aggregation, and that the boundary messaging method recovers some of the information lost when a connected network is partitioned across clients.

The major contributions are summarised as follows:
\begin{itemize}[noitemsep, topsep=0pt]
    \item A client-to-client boundary message-passing mechanism is proposed to exchange spatial-temporal-aware residual signals between adjacent clients without sharing raw traffic observations.
    \item A dual-branch forecasting architecture is introduced to separate shared forecasting structure from client-specific residual corrections. Instead of relying on parameter averaging, clients are coordinated through branch-wise gradient-direction updates: shared-branch gradients are aligned to a population reference, while local-branch gradients are discouraged from collapsing to the same direction.
    \item Extensive experiments are conducted on four real-world benchmark traffic datasets, demonstrating improved forecasting performance with competitive training cost compared with existing federated spatial-temporal forecasting baselines.
\end{itemize}

%------------------------Related Works--------------------------
\section{Related Works}

\subsection{Spatial-Temporal Modelling}
Spatial-temporal modelling aims to capture spatial and temporal dependencies within data, making it central to tasks such as traffic prediction. Early GNNs focused on exploiting static explicit graph structured data \cite{STGCN, DCRNN}. Adaptive approaches later aimed to learn a representation of the graph structure from the data itself, such as AGCRN \cite{AGCRN} which foregoes any graph structure prior and aims to learn a symmetric graph, and Graph Wavenet (GWNet) \cite{GWN} which retains the static graph as a support, and learns bi-directional node embeddings as a correction. 

A parallel line of work applies seasonal-trend decomposition to disentangle periodic, trend, and high-frequency residual components for improved spatial-temporal modelling \cite{DSTG, D2STGNN}. Most recently, Cao et al. \cite{STDN} reframe seasonal-trend decomposition from a spatial-temporal perspective, extracting the trend component conditional on each node's spatial location and temporal context rather than per-node or purely temporally. Such approaches rely on a centralised data setting and do not address distributed real-world systems. 

\subsection{Federated Learning}
FL was developed to enable collaborative client-local training on distributed data \cite{FedAvg}, and has been increasingly explored for spatial-temporal modelling. MFVSTGNN~\cite{MFVSTGNN} introduces a multilevel federated spatial-temporal graph framework that combines local traffic modelling with federated knowledge sharing across clients, while FedGTP \cite{FedGTP} models inter-client spatial dependencies via a polynomial decomposition over client-encoded representations. Furthermore, pFedCTP \cite{pFedCTP} performs personalised knowledge transfer between target areas through adaptive parameter aggregation, and FedDis \cite{FedDis} shares a global traffic pattern bank, while disentangling and capturing local client-specific patterns. These methods collaborate through parameter aggregation on shared components or through the aggregation of learned spatial-temporal representations. However, they do not intervene directly in the underlying optimisation dynamics or exchange targeted, decomposition-aware signals between physically adjacent clients.

A separate line of work intervenes in training dynamics directly; SCAFFOLD \cite{SCAFFOLD} and FedDyn \cite{FedDyn} use gradient correction and dynamic regularisation to align local and global optimisation, and Lu et al. \cite{FedHCA2} propose conflict-averse gradient aggregation across heterogeneous clients. In the federated graph learning literature, FedIGL \cite{FedIGL} addresses graph-level classification across clients holding disjoint graphs from molecular, social, and biological domains, using a bi-gradient regularisation strategy to disentangle invariant from domain-specific substructures. None of these methods coordinate client collaboration through gradient-direction alignment or exchange targeted, decomposition-aware residual signals between physically adjacent clients.

\section{Problem Formulation}

\begin{figure*}[t!]
  \centering
  \includegraphics[width=\textwidth]{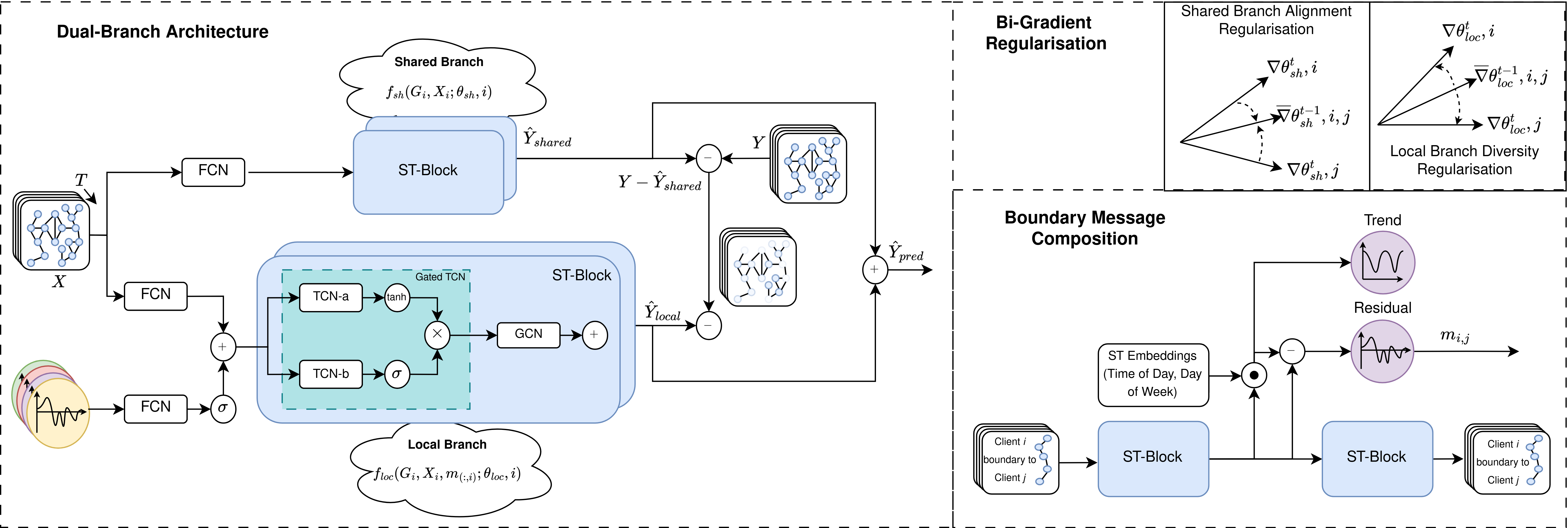}
    \caption{FedeRICo framework. The left panel shows the dual-branch architecture with shared and local branches using the same ST-Block design but separate parameters. The shared branch predicts the main signal, while the local branch predicts residual corrections informed by boundary messages, and both outputs are added for the final forecast. The top-right panel shows shared-gradient alignment and local-gradient diversification, and the bottom-right panel shows sender-local boundary-message composition.}
  \label{fig:FedeRICo_Framework}
\end{figure*}

\textit{\text{\textbf{Definition 1 (Spatial-Temporal Prediction):}} 
Spatial-temporal traffic forecasting predicts future traffic measurements from a window of past observations over a graph network. The network is denoted as $\mathcal{G} = (\mathcal{V}, \mathbf{A})$ with $N = |\mathcal{V}|$ sensors and adjacency $\mathbf{A} \in \mathbb{R}^{N \times N}$ describing pairwise spatial or semantic relationships. Each sensor $v_i$ produces a time series $\mathcal{X}_i = \{x_t\}_{t=1}^{T_{\text{total}}}$, where $x_t \in \mathbb{R}^F$ denotes the traffic state at time $t$. Given a window of $T$ past observations $\mathcal{X}_{t-T+1:t}$, a forecasting model $\mathcal{F}$ produces predictions $\hat{\mathcal{Y}}$ over the next $T'$ steps:
\begin{equation}
\hat{\mathcal{Y}} = \mathcal{F}(\mathcal{X}_{t-T+1:t}, \mathbf{A}; \theta),
\end{equation}
where $\mathcal{F}(\cdot)$ is a learnable model with parameters $\theta$ and $\hat{\mathcal{Y}} \in \mathbb{R}^{N \times T' \times F}$.
}

\noindent \textit{\text{\textbf{Definition 2 (Federated Spatial-Temporal Prediction):}} 
In the federated setting, the sensor set is split across $M$ clients. Client $m$ holds a local subgraph $\mathcal{G}^{(m)} = (\mathcal{V}^{(m)}, \mathbf{A}^{(m)})$ together with its measurements $\mathcal{X}^{(m)}$, where the client partitions are disjoint, $\mathcal{V}^{(m)} \cap \mathcal{V}^{(n)} = \emptyset$ for $m \neq n$, and jointly cover the full network. This partitioning severs the boundary edges $\mathcal{E}_{m,n}$ that originally connected adjacent client subgraphs in the original network, removing a source of spatial information that traditional federated approaches do not recover.
The aim is for clients to collaboratively improve their forecasting performance without aggregating their raw observations. This is formulated as a joint minimisation over per-client parameters:
\begin{equation}
\min_{\{\theta^{(m)}\}_{m=1}^{M}} \sum_{m=1}^M \mathcal{L} \left( \mathcal{F}^{(m)} \left( \mathcal{X}_{t-T+1:t}^{(m)}; \theta^{(m)} \right), \mathcal{Y}^{(m)}  \right)
\end{equation}
\noindent where $\mathcal{L}(\cdot)$ is the local forecasting loss, $\mathcal{Y}^{(m)}$ denotes the future traffic states at client $m$, and each $\theta^{(m)}$ remains local to its client. Our framework aligns client collaboration through complementary mechanisms that operate on gradients and on boundary information, as defined in Section \ref{sec:methodology}.
}

\section{Methodology}
\label{sec:methodology}

Figure \ref{fig:FedeRICo_Framework} provides an overview of the proposed FedeRICo framework. This approach is designed for federated spatial-temporal forecasting under graph partitioning, where each client owns a different road subgraph and therefore learns spatial operators that are not directly interchangeable. The framework addresses two complementary challenges. First, clients should benefit from each other's optimisation dynamics without requiring full parameter aggregation across heterogeneous graphs. Second, adjacent clients should be able to exchange limited boundary information so that disturbances across partition boundaries are not entirely removed by the federated split. 

This section describes the components in the framework which address these challenges. Section \ref{sec:dual_branch_architecture} introduces the dual-branch forecasting model and its residual training objective. Section \ref{sec:client_similarity_gradient} describes the gradient alignment and divergence regularisation method. Section \ref{sec:boundary_message_composition} presents the construction and injection of boundary residual messages.

\subsection{Dual-Branch Architecture}
\label{sec:dual_branch_architecture}

FedeRICo uses a dual-branch forecasting architecture in which both branches are built from spatial-temporal (ST) Blocks based on Graph WaveNet~\cite{GWN}. Each ST-Block combines gated temporal convolutions with graph diffusion convolution. For client $m$, let $N_m = |\mathcal{V}^{(m)}|$. The input traffic window $\mathcal{X}^{(m)}_{t-T+1:t}$ is first projected to a hidden representation. Denote the input to the $\ell$-th ST-Block by $\mathbf{H}^{(m)}_{\ell}$. The temporal module applies a gated temporal convolution:

\begin{gather}
\mathbf{U}^{(m)}_{\ell} = \tanh\!\left( \Theta^{f}_{\ell} \star \mathbf{H}^{(m)}_{\ell} \right), \nonumber \\
\mathbf{R}^{(m)}_{\ell} = \sigma\!\left( \Theta^{g}_{\ell} \star \mathbf{H}^{(m)}_{\ell} \right), \nonumber \\
\mathbf{Z}^{(m)}_{\ell} = \mathbf{U}^{(m)}_{\ell} \odot \mathbf{R}^{(m)}_{\ell}. \label{eq:gated_mechanism}
\end{gather}
where $\mathbf{U}^{(m)}_{\ell}$ is the candidate temporal response, $\mathbf{R}^{(m)}_{\ell}$ is the temporal gate, $\Theta^{f}_{\ell}$ and $\Theta^{g}_{\ell}$ are the filter and gate temporal convolution kernels, $\star$ denotes temporal convolution, $\sigma(\cdot)$ is the sigmoid function, and $\odot$ denotes element-wise multiplication.

Spatial dependencies are then modelled through graph diffusion convolution over physical and learned graph supports. Following bi-directional diffusion convolution~\cite{DCRNN,GWN}, the physical supports are the forward random-walk transition matrix induced by $\mathbf{A}^{(m)}$ and the reverse transition matrix induced by ${\mathbf{A}^{(m)}}^{\top}$. In addition, each client learns an adaptive graph support from trainable node embeddings:
\begin{equation}
\widehat{\mathbf{A}}^{(m)}_{\mathrm{adp}} = \operatorname{softmax} \left( \operatorname{ReLU} \left( \mathbf{E}^{(m)}_1 \mathbf{E}^{(m)}_2 \right) \right),
\end{equation}
where $\mathbf{E}^{(m)}_1 \in \mathbb{R}^{N_m \times d}$ and $\mathbf{E}^{(m)}_2 \in \mathbb{R}^{d \times N_m}$ are learnable client-specific node embeddings, and the softmax is applied row-wise. Because this support is produced from two unconstrained embedding matrices it can represent asymmetric spatial influence. This is compatible with boundary-message inputs since information introduced at boundary nodes can be propagated through a directed spatial operator. The resulting adaptive support is used together with the physical supports, so each client's spatial propagation depends on both the observed road topology and a learned client-specific graph operator. 

The shared and local branches use the same ST-Block backbone but maintain separate parameters. The shared branch receives only the local traffic observations, while the local branch receives both the local observations and the boundary message representation from adjacent clients. Boundary messages are projected to the model hidden dimension and added to the local branch hidden representation through a learned gate, after the initial input projection. The final prediction is composed additively:
\begin{equation}
\widehat{\mathcal{Y}}^{(m)} = \widehat{\mathcal{Y}}^{(m)}_{\mathrm{sh}} + \widehat{\mathcal{Y}}^{(m)}_{\mathrm{loc}},
\end{equation}
where $\widehat{\mathcal{Y}}^{(m)}_{\mathrm{sh}}$ and $\widehat{\mathcal{Y}}^{(m)}_{\mathrm{loc}}$ denote the shared-branch and local-branch outputs, respectively.

During training, the shared branch is supervised against the forecasting target, while the local branch is supervised against the residual left by the shared branch. This decomposition encourages the shared branch to capture the dominant local forecasting signal, while the local branch focuses on residual corrections informed by boundary messages.

\subsection{Bi-Gradient Regularisation}
\label{sec:client_similarity_gradient}

% Clients sharing the same model architecture still have differ underlying spatial operators tied to different node sets, physical supports, and learned adaptive adjacencies. FedeRICo avoids direct averaging of spatial-temporal parameters and instead cooridnates clients through gradient direction. 

% Client compatibility is estimated from the graph operators learned by the forecasting backbone. In GWNet \cite{GWN} each client learns an adaptive graph support to supplement the physical road network. FedeRICo constructs structural signatures from the resulting client graph operators and compares clients using GraphWave \cite{graphwave}, which represents nodes through heat-kernel diffusion responses, capturing structural roles rather than node identities. Since clients contain different node sets, the resulting node-signature distributions are compared with sliced Wasserstein-distance \cite{wasserstein_slice}. A sparse mutual-neighbour rule over this distance matrix defines the compatible client set $\mathcal{C}(m)$ for each client $m$. 

FedeRICo coordinates clients through their gradients rather than their parameters. As shown in Table~\ref{tab:all_datasets_results}, parameter averaging performs poorly under heterogeneous client partitions, suggesting that direct aggregation can dilute client-specific representations. This approach instead draws on the bi-gradient principle of FedIGL~\cite{FedIGL}, in which gradients associated with a shared branch are encouraged to align across clients, while gradients associated with a client-specific branch are discouraged from collapsing to the same direction.

This framework adapts this principle to spatial-temporal forecasting by applying coordination directly to gradient directions after backpropagation and before the optimiser step, preserving each client's gradient norm. Both alignment and diversification operate on cosine similarity rather than $L_2$ distance, since gradient magnitudes can vary significantly across clients due to different underlying spatial operators. Cosine similarity isolates directional agreement, providing a coordination signal that is more robust to magnitude differences that may arise from heterogeneous graph spectra and local traffic dynamics~\cite{FedFV, CFL}.

Let $\mathbf{g}^{(m)}_{\mathrm{sh}}$ and $\mathbf{g}^{(m)}_{\mathrm{loc}}$ denote the current gradients of the shared and local branches of client $m$ for non-node-specific parameters. At the end of each communication round, FedeRICo estimates all-client reference directions from the accumulated client gradients. The shared-branch reference for client $m$ is denoted $\bar{\mathbf{g}}^{(m)}_{\mathrm{sh}}$, and the local-branch reference is denoted $\bar{\mathbf{g}}^{(m)}_{\mathrm{loc}}$.

Let $\mathbf{u}(\mathbf{g})=\mathbf{g}/\|\mathbf{g}\|_2$ denote the unit direction of a non-zero gradient. For the shared branch, the client gradient direction is merged with the population reference direction,
\begin{gather}
\mathbf{a}^{(m)}_{\mathrm{sh}} = (1-\lambda_{\mathrm{align}}) \mathbf{u}\left(\mathbf{g}^{(m)}_{\mathrm{sh}}\right) + \lambda_{\mathrm{align}} \mathbf{u}\left(\bar{\mathbf{g}}^{(m)}_{\mathrm{sh}}\right), \nonumber \\
\widetilde{\mathbf{g}}^{(m)}_{\mathrm{sh}} = \|\mathbf{g}^{(m)}_{\mathrm{sh}}\|_2 \mathbf{u}\left(\mathbf{a}^{(m)}_{\mathrm{sh}}\right). \label{eq:grad_alignment}
\end{gather}

For the local branch, this approach only modifies the gradient when it is too aligned with the population local-branch reference:
\begin{gather}
c_m = \cos\left( \mathbf{g}^{(m)}_{\mathrm{loc}}, \bar{\mathbf{g}}^{(m)}_{\mathrm{loc}} \right), \nonumber \\
\mathbf{a}^{(m)}_{\mathrm{loc}} = \mathbf{u}\left(\mathbf{g}^{(m)}_{\mathrm{loc}}\right) - \lambda_{\mathrm{div}}[c_m-\delta]_+ \mathbf{u}\left(\bar{\mathbf{g}}^{(m)}_{\mathrm{loc}}\right), \nonumber \\
\widetilde{\mathbf{g}}^{(m)}_{\mathrm{loc}} = \|\mathbf{g}^{(m)}_{\mathrm{loc}}\|_2 \mathbf{u}\left(\mathbf{a}^{(m)}_{\mathrm{loc}}\right). \label{eq:grad_diversification}
\end{gather}

\noindent where $[x]_+=\max(x,0)$ denotes the positive-part operator, $\lambda_{\mathrm{div}}$ controls the strength of the diversification update, and $\delta$ is the local-branch diversity margin.

The modified gradients are then used for the optimiser step. FedeRICo therefore couples clients through branch-level gradient directions rather than through direct parameter averaging. Shared branches are guided toward population-level forecasting structure, while the local branch remains free to preserve personalised residual corrections, including those induced by boundary messages.

\subsection{Boundary Message Composition}
\label{sec:boundary_message_composition}

Federated partitioning removes physical edges that originally connected neighbouring road regions. These removed edges are important for traffic forecasting because congestion, incidents, and other short-term disturbances often propagate across the partition boundary. The proposed framework therefore introduces boundary message composition to integrate an encoded view of cross-client spatial context.

This approach does not communicate full raw trajectories or low-frequency traffic structure; such signals may encode persistent demand levels, commuting periodicity, and other commercially sensitive regional patterns. Instead, the communicated message is designed to represent boundary-local residual behaviour, such as non-periodic short-term deviations near the boundary after smooth temporal structure has been suppressed.

For a neighbouring client pair $(n,m)$, let $\mathcal{B}_{m \leftarrow n}$ denote the receiver-side boundary nodes in client $m$ adjacent to client $n$. For each receiver boundary node, FedeRICo selects the top-$k$ sender nodes in client $n$ closest to the cross-client interface. The sender-side boundary packet is
\begin{equation}
    \mathbf{B}^{(u,t)}_{n \rightarrow m}
    =
    \left[
    \mathbf{x}^{(v_1)}_{t-W+1:t},
    \ldots,
    \mathbf{x}^{(v_k)}_{t-W+1:t}
    \right],
\end{equation}
\noindent where $u \in \mathcal{B}_{m \leftarrow n}$ is the receiver boundary node, $\{v_1,\ldots,v_k\}$ are the selected sender nodes, and $W$ is the boundary history window. 

The packet is processed by a lightweight boundary spatial-temporal module. This module first encodes the sender boundary packet using gated temporal convolutions and a learned mixing operator over the selected sender nodes. Let $\mathbf{Z}^{(u,t)}_{n \rightarrow m}$ denote the encoded boundary representation produced by this first boundary ST module. Following the motivation of spatial-temporal decomposition methods such as STDN~\cite{STDN}, this approach uses temporal embeddings such as time-of-day and day-of-week to estimate and suppress periodic structure in the encoded boundary representation:
\begin{equation}
    \mathbf{R}^{(u,t)}_{n \rightarrow m}
    =
    \mathbf{Z}^{(u,t)}_{n \rightarrow m}
    -
    \mathbf{Z}^{(u,t)}_{n \rightarrow m}
    \odot
    \sigma\!\left(q_{\psi}(\mathbf{T}_t)\right),
\end{equation}
\noindent where $\mathbf{T}_t$ denotes the temporal embedding, $q_{\psi}(\cdot)$ projects the temporal embedding to the boundary representation dimension, $\sigma(\cdot)$ is the sigmoid function, and $\odot$ denotes element-wise multiplication. The resulting residual representation $\mathbf{R}^{(u,t)}_{n \rightarrow m}$ suppresses time-periodic structure while retaining short-term boundary deviations.

A second boundary ST module processes this residual representation to form the transmitted message:
\begin{equation}
    \mathbf{m}^{(u,t)}_{n \rightarrow m}
    =
    h_{\phi}
    \left(
    \mathbf{R}^{(u,t)}_{n \rightarrow m}
    \right),
\end{equation}
\noindent where $h_{\phi}(\cdot)$ denotes the residual boundary-message module. The output $\mathbf{m}^{(u,t)}_{n \rightarrow m}$ is therefore a learned residual boundary message rather than a raw boundary trajectory.

For client $m$, all incoming messages from adjacent clients are assembled into a boundary context tensor $\mathbf{M}^{(m)}$. This context is provided only to the local branch, together with the client's own input window and graph supports. The shared branch does not receive boundary messages. This confines cross-client information to the personalised residual-correction pathway while keeping the shared branch focused on the client's own spatial-temporal structure.

\subsection{Federated Framework}

The proposed framework has two stages. First, each client trains a small sender-local boundary-message encoder on its own boundary nodes, where boundary nodes are determined by cross-client proximity. This pretraining uses only local sender-side traffic input data $\mathbf{B}^{(u,t)}_{n \rightarrow m}$, and sender-side forecasting targets. Hence, there is no cross-client communication or collaboration in pre-training this client-specific module. In our final configuration, these encoders are trained for 20 local epochs and then frozen, so during federated training they produce compact boundary messages for neighbouring receiver clients. 

In the main federated stage, each client maintains a personalised spatial-temporal model with shared and local branches. At each communication round, clients perform local optimisation on their own subgraphs. For each batch, incoming boundary messages are constructed from adjacent sender clients, aligned by time, and injected into the receiver model. The server coordinates learning using normalised gradients from model branches. Shared branches are encouraged to follow the reference directions, while private branches are discouraged from collapsing to the same direction. Thus, clients exchange model-update information through server and boundary-context messages across adjacent clients, but raw node observations remain local. Clients do not transmit raw traffic observations, boundary messages, or per-batch gradients to the server. Each client only communicates round-level branch gradients used to form reference directions for the next communication round. The server-side mechanism is detailed in Algorithm \ref{alg:FedeRICo_training}.

\algrenewcommand\algorithmicindent{0.8em}
\begin{algorithm}[t]
\caption{FedeRICo Federated Training}
\label{alg:FedeRICo_training}
\begin{algorithmic}[1]
\Require Clients $\mathcal{C}$, datasets $\{\mathcal{D}_k\}$, boundary encoders $\{\psi_k\}$
\Require Rounds $R$, local epochs $E$
\Ensure Personalised client models $\{\theta_k\}$

\State Initialise client models $\{\theta_k\}$.
\State Set server reference directions as unavailable.

\For{federated round $r = 1,\dots,R$}
    \If{$r > 1$}
        \State Server sends each client its reference directions.
    \EndIf

    \For{each client $k$ in parallel}
        \State Initialise branch-gradient accumulators.

        \For{local epoch $e = 1,\dots,E$}
            \For{each mini-batch}
                \State Encode packets $\mathbf{B}^{(u,t)}_{n \rightarrow k}$ into context $C_k(t)$.
                \State Predict $\hat{Y}_k = f_{\theta_k}(X_k,C_k(t))$.
                \State Compute $\mathcal{L}_k$ using shared and local-residual targets.
                \State Backpropagate $\nabla_{\theta_k}\mathcal{L}_k$.
                \State Accumulate shared and local branch gradients.

                \If{$r > 1$}
                    \State Align shared gradients to reference from Eq. (\ref{eq:grad_alignment}).
                    \State Diverge local gradients from reference from Eq. (\ref{eq:grad_diversification}).
                \EndIf

                \State Update $\theta_k$.
            \EndFor
        \EndFor

        \State Send round-level gradient summaries to the server.
    \EndFor

    \State Server forms references for the next round.
    \State Validate clients and retain best checkpoints.
\EndFor
\end{algorithmic}
\end{algorithm}

\section{Experiments}

\subsection{Datasets}

\begin{table}[ht]
\centering
\caption{Statistics of traffic datasets.}
\label{tab:metrla_4_clients_voronoi}
\begin{tabular}{ccccc}
\hline
\textbf{Datasets} & \textbf{\# Sensors} & \textbf{\# Samples} & \textbf{Target Feature} \\
\hline
METR-LA & 207 & 34,272 & Speed \\
PEMS-BAY & 325 & 52,116 & Speed \\
PEMS03 & 358 & 26,208 & Traffic Flow \\
PEMS07 (M) & 228 & 12,672 & Speed\\
\hline
\end{tabular}
\end{table}

\noindent FedeRICo is evaluated on four real-world traffic forecasting datasets commonly used in spatial-temporal prediction: METR-LA, PEMS-BAY \cite{DCRNN}, PEMS03 \cite{STSGCN}, and PEMS07 (M) \cite{STGCN}. All datasets record data at 5-minute intervals. METR-LA and PEMS-BAY contain traffic speed measurements from freeway sensors in the Los Angeles County and the Bay Area, respectively. PEMS03 and PEMS07 (M) are collected from California Performance Measurement System (PEMS) districts, with PEMS03 using traffic flow and PEMS07 (M) using traffic speed as the prediction target.

\subsection{Baseline Methods}
To assess the performance of FedeRICo, several baseline centralised spatial-temporal forecasting models and federated traffic forecasting methods are used. The centralised models are trained with centralised, full road network datasets to provide reference performance for the non-federated setting. The federated baselines operate under the same partitioning protocol as FedeRICo.

\begin{itemize}[leftmargin=*, itemsep=0.2em, topsep=0.2em]
    \item \textbf{STDN} \cite{STDN}: Decomposes traffic signals into trend and seasonality components via spatial-temporal embeddings, encoding each stream with separate GRUs before decoding with multi-head attention. 
    \item \textbf{GWNet} \cite{GWN}: Combines stacked layers of dilated temporal convolutions with graph convolutions over both predefined and learned adaptive graph supports.
    \item \textbf{AGCRN} \cite{AGCRN}: Learns node-specific embeddings to construct an adaptive graph for recurrent spatial-temporal modelling.
\end{itemize} 

\noindent For the federated traffic prediction methods, the following baselines are implemented:
\begin{itemize}[leftmargin=*, itemsep=0.2em, topsep=0.2em]
    \item \textbf{FedAvg} \cite{FedAvg}: Trains local client models and periodically averages model parameters across clients using GWNet as the spatial-temporal model.
    \item \textbf{MFVSTGNN}~\cite{MFVSTGNN}: Federates a multi-view spatial-temporal graph neural network, using separate local and collaborative views to model client-specific traffic dynamics and shared spatial-temporal dependencies.
    \item \textbf{FedGTP} \cite{FedGTP}: Performs federated traffic prediction on an AGCRN-based model by exchanging learned spatial-temporal representations between clients through a graph-based collaborative training mechanism.
    \item \textbf{pFedCTP} \cite{pFedCTP}: Coordinates FL through a shared temporal module and a private spatial module, with personalisation achieved by adaptively aggregating shared parameters.
    \item \textbf{FedDis} \cite{FedDis}: Implements an AGCRN-based spatial temporal model with separate shared and personalised representations using client similarity to guide federated parameter aggregation.
\end{itemize}

\subsection{Experiment Settings}
All experiments are implemented in PyTorch and conducted on NVIDIA RTX 4090 GPUs. Each dataset is split into training, validation, and test sets using a 60:20:20 ratio. Inputs are processed into $T=12$ window time steps, with the prediction horizon $T'=12$. All models are trained using the Adam optimiser \cite{adam} with a learning rate of $1e^{-3}$, and batch size 128. Federated methods are trained for 100 global rounds and 1 local round of training. Source code is made available at \url{https://anonymous.4open.science/r/FedeRICo-CA6B}.

\section{Results}

\subsection{Performance Comparison}

\begin{table*}[t]
\centering
\small
\caption{Results across METR-LA, PEMS-BAY, PEMS-D3, and PEMS-D7 datasets.}
\label{tab:all_datasets_results}
\resizebox{\textwidth}{!}{%
\begin{tabular}{cc ccc ccc ccc ccc}
\toprule
\textbf{Framework} & \textbf{Model}
& \multicolumn{3}{c}{\textbf{METR-LA}}
& \multicolumn{3}{c}{\textbf{PEMS-BAY}}
& \multicolumn{3}{c}{\textbf{PEMS-D3}}
& \multicolumn{3}{c}{\textbf{PEMS-D7}} \\
\cmidrule(lr){3-5}
\cmidrule(lr){6-8}
\cmidrule(lr){9-11}
\cmidrule(lr){12-14}
\textbf{} & \textbf{}
& \textbf{MAE} & \textbf{RMSE} & \textbf{MAPE}
& \textbf{MAE} & \textbf{RMSE} & \textbf{MAPE}
& \textbf{MAE} & \textbf{RMSE} & \textbf{MAPE}
& \textbf{MAE} & \textbf{RMSE} & \textbf{MAPE} \\
\midrule
Centralised & STDN & 3.01 & 6.07 & 8.12 & 1.65 & 3.74 & 3.69 & 15.04 & 26.36 & 15.20 & 2.55 & 5.09 & 6.36 \\
Centralised & GWNet & 3.09 & 6.20 & 8.38 & 1.61 & 3.58 & 3.62 & 14.81 & 25.38 & 14.74 & 2.60 & 5.09 & 6.61 \\
Centralised & AGCRN & 3.26 & 6.52 & 9.06 & 1.67 & 3.79 & 3.80 & 15.90 & 27.46 & 16.11 & 2.77 & 5.34 & 6.94 \\
\midrule
Federated   & FedAvg  & 3.93 & 7.49 & 11.98 & 1.86 & 4.04 & 4.30 & 21.07 & 33.48 & 25.44 & 3.03 & 5.71 & 7.82 \\
Federated   & FedGTP & 3.92 & 7.61 & 11.82 & 1.97 & 4.37 & 4.62 & 18.49 & 30.38 & 20.05 & 3.01 & 5.93 & 7.79 \\
Federated   & MFVSTGNN  & 3.79 & 7.43 & 10.82 & 1.83 & 4.09 & 4.21 & 17.30 & 28.88 & 17.64 & 3.10 & 5.89 & 7.95 \\
Federated   & pFedCTP & 3.76 & 7.01 & 10.39 & 2.05 & 4.61 & 4.88 & 20.83 & 33.85 & 22.68 & 3.38 & 6.33 & 8.71 \\
Federated   & FedDis  & 3.51 & 6.98 & 10.02 & 1.79 & 3.95 & 4.09 & 16.54 & 28.43 & 16.42 & 2.97 & 5.70 & 7.42 \\
\midrule
Federated   & FedeRICo & 3.19 & 6.31 & 8.87 & 1.67 & 3.67 & 3.72 & 15.89 & 26.83 & 16.03 & 2.71 & 5.24 & 6.76 \\
\bottomrule
\end{tabular}%
}
\end{table*}

Table~\ref{tab:all_datasets_results} reports the forecasting performance across four traffic benchmarks under centralised and federated settings. FedeRICo achieves the best federated performance on every dataset and metric, consistently improving over both parameter-averaging baselines and recent personalised and collaborative FL methods. Compared with FedAvg, FedeRICo reduces MAE by $18.8\%$ on METR-LA, $10.2\%$ on PEMS-BAY, $24.6\%$ on PEMS-D3, and $10.6\%$ on PEMS-D7. 

The comparison also highlights a gap between existing federated traffic prediction methods and centralised methods. The performance of FedGTP, MFVSTGNN, pFedCTP, and FedDis remain substantially below all centralised models on most datasets, suggesting that direct parameter averaging or representation-level collaboration may be insufficient when traffic networks are partitioned into heterogeneous client subgraphs. FedeRICo narrows this gap most consistently: its MAE remains within $3.2\% - 7.3\%$  of centralised GWNet across datasets. Moreover, FedeRICo matches or outperforms centralised AGCRN across all datasets, despite operating without centralising client traffic datasets.

Relative to the strongest federated baseline, FedDis, FedeRICo lowers MAE by $0.32$ on METR-LA ($9.1\%$), $0.12$ on PEMS-BAY ($6.7\%$), $0.65$ on PEMS-D3 ($3.9\%$), and $0.26$ on PEMS-D7 ($8.8\%$). These results support the main design motivation of FedeRICo. Boundary messages provide cross-client spatial context without merging raw subgraphs, while branch-gradient coordination avoids the detrimental effect caused by averaging parameters.

\subsection{Ablation Studies}

\subsubsection{FL Framework Ablation}

\begin{figure*}[t]
\centering
\scriptsize

% =========================
% Manual legend
% Color-blind-friendly Okabe-Ito palette
% =========================
\begin{tikzpicture}
    \node[draw=black, fill=black!30, minimum width=0.30cm, minimum height=0.12cm, inner sep=0pt] at (0,0) {};
    \node[right, font=\scriptsize] at (0.22,0) {w/o Coll};

    \node[draw=black, fill={rgb,255:red,86;green,180;blue,233}, minimum width=0.30cm, minimum height=0.12cm, inner sep=0pt] at (1.70,0) {};
    \node[right, font=\scriptsize] at (1.92,0) {w/o BM};

    \node[draw=black, fill={rgb,255:red,213;green,94;blue,0}, minimum width=0.30cm, minimum height=0.12cm, inner sep=0pt] at (3.20,0) {};
    \node[right, font=\scriptsize] at (3.42,0) {w/o GA};

    \node[draw=black, fill={rgb,255:red,0;green,158;blue,115}, minimum width=0.30cm, minimum height=0.12cm, inner sep=0pt] at (4.70,0) {};
    \node[right, font=\scriptsize] at (4.92,0) {W OS};

    \node[draw=black, fill={rgb,255:red,230;green,159;blue,0}, minimum width=0.30cm, minimum height=0.12cm, inner sep=0pt] at (6.05,0) {};
    \node[right, font=\scriptsize] at (6.27,0) {FedeRICo};
\end{tikzpicture}

\vspace{0.08cm}

\resizebox{\textwidth}{!}{%
\begin{tikzpicture}
    \begin{groupplot}[
        group style={
            group size=6 by 1,
            horizontal sep=0.62cm
        },
        width=2.85cm,
        height=3.05cm,
        ybar,
        /pgf/bar width=5.2pt,
        /pgf/bar shift=0pt,
        xmin=0.5,
        xmax=5.5,
        enlarge x limits=false,
        clip=false,
        xtick=\empty,
        axis lines=box,
        axis line style={line width=0.25pt},
        grid=both,
        minor tick num=1,
        major grid style={gray!50, dash pattern=on 0.8pt off 0.8pt, line width=0.16pt},
        minor grid style={gray!28, dash pattern=on 0.6pt off 0.6pt, line width=0.12pt},
        tick align=outside,
        tick pos=left,
        tick style={black, line width=0.1pt},
        yticklabel style={font=\fontsize{3.8}{3.1}\selectfont, xshift=1pt},
        ylabel style={font=\fontsize{4.0}{3.3}\selectfont, yshift=-0.1cm},
        label style={font=\fontsize{4.0}{3.3}\selectfont},
        title style={at={(0.5,-0.30)}, anchor=north, font=\fontsize{4.2}{3.5}\selectfont\bfseries\itshape}
    ]

    % =========================
    % (a) METR-LA MAE
    % =========================
    \nextgroupplot[
        ylabel={MAE},
        ymin=3.10, ymax=3.52,
        ytick={3.10,3.20,3.30,3.40,3.50},
        title={(a)}
    ]
    \addplot[fill=black!30, draw=black] coordinates {(1,3.3467)};
    \addplot[fill={rgb,255:red,86;green,180;blue,233}, draw=black] coordinates {(2,3.2845)};
    \addplot[fill={rgb,255:red,213;green,94;blue,0}, draw=black] coordinates {(3,3.4856)};
    \addplot[fill={rgb,255:red,0;green,158;blue,115}, draw=black] coordinates {(4,3.1966)};
    \addplot[fill={rgb,255:red,230;green,159;blue,0}, draw=black] coordinates {(5,3.1878)};
    
    % =========================
    % (b) METR-LA RMSE
    % =========================
    \nextgroupplot[
        ylabel={RMSE},
        ymin=6.15, ymax=6.75,
        ytick={6.15,6.25,6.35,6.45,6.55,6.65,6.75},
        title={(b)}
    ]
    \addplot[fill=black!30, draw=black] coordinates {(1,6.7082)};
    \addplot[fill={rgb,255:red,86;green,180;blue,233}, draw=black] coordinates {(2,6.5712)};
    \addplot[fill={rgb,255:red,213;green,94;blue,0}, draw=black] coordinates {(3,6.6066)};
    \addplot[fill={rgb,255:red,0;green,158;blue,115}, draw=black] coordinates {(4,6.3200)};
    \addplot[fill={rgb,255:red,230;green,159;blue,0}, draw=black] coordinates {(5,6.3095)};
    
    % =========================
    % (c) METR-LA MAPE
    % =========================
    \nextgroupplot[
          ylabel={MAPE},
          ymin=8.00, ymax=10.90,
          ytick={8.0,8.4,8.8,9.2,9.6,10.0,10.4,10.8},
          title={(c)}
      ]
    \addplot[fill=black!30, draw=black] coordinates {(1,9.16)};
    \addplot[fill={rgb,255:red,86;green,180;blue,233}, draw=black] coordinates {(2,9.14)};
    \addplot[fill={rgb,255:red,213;green,94;blue,0}, draw=black] coordinates {(3,10.76)};
    \addplot[fill={rgb,255:red,0;green,158;blue,115}, draw=black] coordinates {(4,8.84)};
    \addplot[fill={rgb,255:red,230;green,159;blue,0}, draw=black] coordinates {(5,8.87)};
    
    % =========================
    % (d) PEMS-BAY MAE
    % =========================
    \nextgroupplot[
        ylabel={MAE},
        ymin=1.64, ymax=1.74,
        ytick={1.64,1.66,1.68,1.70,1.72,1.74},
        title={(d)}
    ]
    \addplot[fill=black!30, draw=black] coordinates {(1,1.7177)};
    \addplot[fill={rgb,255:red,86;green,180;blue,233}, draw=black] coordinates {(2,1.6943)};
    \addplot[fill={rgb,255:red,213;green,94;blue,0}, draw=black] coordinates {(3,1.6834)};
    \addplot[fill={rgb,255:red,0;green,158;blue,115}, draw=black] coordinates {(4,1.6764)};
    \addplot[fill={rgb,255:red,230;green,159;blue,0}, draw=black] coordinates {(5,1.6680)};
    
    % =========================
    % (e) PEMS-BAY RMSE
    % =========================
    \nextgroupplot[
        ylabel={RMSE},
        ymin=3.62, ymax=3.82,
        ytick={3.62,3.66,3.70,3.74,3.78,3.82},
        title={(e)}
        ]
    \addplot[fill=black!30, draw=black] coordinates {(1,3.7900)};
    \addplot[fill={rgb,255:red,86;green,180;blue,233}, draw=black] coordinates {(2,3.7752)};
    \addplot[fill={rgb,255:red,213;green,94;blue,0}, draw=black] coordinates {(3,3.6942)};
    \addplot[fill={rgb,255:red,0;green,158;blue,115}, draw=black] coordinates {(4,3.6979)};
    \addplot[fill={rgb,255:red,230;green,159;blue,0}, draw=black] coordinates {(5,3.6713)};
    
    % =========================
    % (f) PEMS-BAY MAPE
    % =========================
    \nextgroupplot[
        ylabel={MAPE},
        ymin=3.68, ymax=3.84,
        ytick={3.68,3.72,3.76,3.80,3.84},
        title={(f)}
    ]
    \addplot[fill=black!30, draw=black] coordinates {(1,3.81)};
    \addplot[fill={rgb,255:red,86;green,180;blue,233}, draw=black] coordinates {(2,3.81)};
    \addplot[fill={rgb,255:red,213;green,94;blue,0}, draw=black] coordinates {(3,3.82)};
    \addplot[fill={rgb,255:red,0;green,158;blue,115}, draw=black] coordinates {(4,3.76)};
    \addplot[fill={rgb,255:red,230;green,159;blue,0}, draw=black] coordinates {(5,3.72)};
    \end{groupplot}
\end{tikzpicture}%
}

\vspace{0.05cm}

\caption{Ablation study. Subfigures (a), (b), and (c) are the results of the METR-LA ablation experiments, while subfigures (d), (e), and (f) present the results of the PEMS-BAY ablation experiments.}
\label{fig:ablation_study}
\end{figure*}
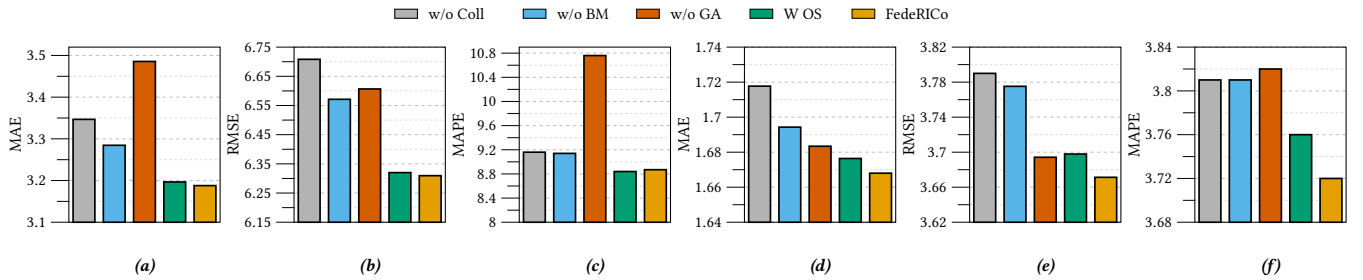

Figure~\ref{fig:ablation_study} isolates the contribution of each component in FedeRICo. \textbf{w/o Coll} trains the dual-branch model locally without federated collaboration, \textbf{w/o BM} removes cross-client boundary messages while retaining gradient alignment, \textbf{w/o GA} replaces gradient alignment with standard parameter averaging, \textbf{w OS} uses operator-similarity-based client selection, and \textbf{FedeRICo} denotes the full proposed framework.

The local-only dual-branch variant (\textbf{w/o Coll}) already provides a competitive personalised forecasting model, but it is consistently worse than the collaborative variants on both METR-LA and PEMS-BAY. This shows that the gains of this approach do not come solely from increasing model capacity through the dual-branch design; cross-client coordination is necessary to recover spatial dependencies that are broken by graph partitioning.

Removing boundary messages (\textbf{w/o BM}) degrades performance relative to the full model across both datasets. On METR-LA, MAE increases from $3.1878$ to $3.2845$, while RMSE increases from $6.3095$ to $6.5712$. These drops indicate that boundary messages provide useful cross-client residual context beyond what can be recovered through gradient coordination alone.

The \textbf{w/o GA} variant evaluates whether standard federated parameter averaging can replace the proposed branch-gradient coordination. This ablation is particularly revealing on METR-LA, where MAE rises to $3.4856$ and MAPE increases sharply to $10.76$. The degradation supports the premise that directly averaging parameters across heterogeneous traffic subgraphs can interfere with client-specific representations. Gradient-space coordination provides a softer coupling mechanism, aligning the shared branch while preserving local residual specialisation.

For the \textbf{W OS} variant, client compatibility is estimated from the graph operators learned by the forecasting backbone. In GWNet~\cite{GWN}, each client learns an adaptive graph support in addition to the physical road network. We construct structural signatures from these learned graph operators using GraphWave~\cite{graphwave}, which characterises nodes by heat-kernel diffusion responses and therefore captures structural roles rather than node identities. Since clients contain disjoint node sets, we compare the resulting signature distributions using sliced Wasserstein distance~\cite{wasserstein_slice}. The operator-similarity variant (\textbf{W OS}) performs competitively, but does not consistently improve over all-client coordination. This suggests that the learned gradient references are sufficiently stable to benefit from broader all-client coordination.

Overall, the ablation results show that the two main mechanisms are complementary. Boundary messages restore missing spatial context at partition boundaries, while branch-gradient coordination prevents this collaboration from collapsing into destructive parameter averaging.

\subsubsection{Hyperparameter Ablation}

We study the sensitivity of FedeRICo to two groups of hyperparameters on METR-LA: boundary-message configuration and gradient-controller strength. For boundary messages, we vary the message hidden dimension $d_m$, the temporal decorrelation coefficient $\lambda_t$, and the number of received sender contexts top-$k$. For the gradient controller, we jointly vary $\lambda_{\mathrm{align}}$, $\lambda_{\mathrm{div}}$, and $\delta$ using a shared value, testing whether the strength of branch-gradient coordination requires narrow tuning.

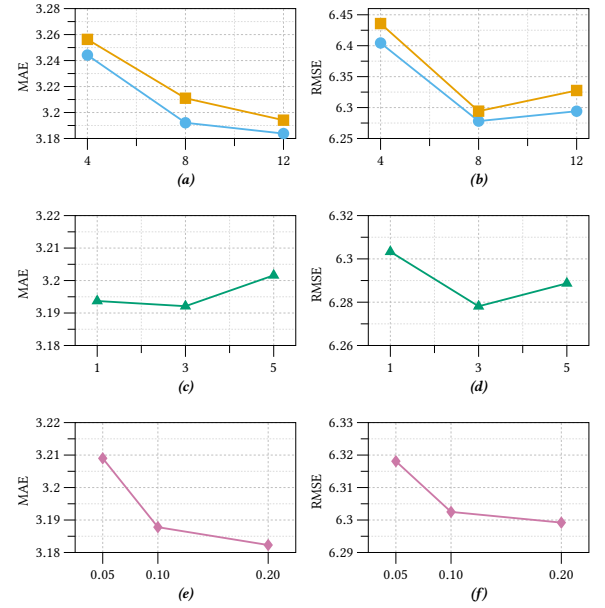
\begin{figure}[t]
\centering
\scriptsize
\begin{tikzpicture}
\begin{groupplot}[
        group style={
            group size=2 by 3,
            horizontal sep=0.96cm,
            vertical sep=1.02cm
        },
        width=4.50cm,
        height=3.30cm,
        axis lines=box,
        axis line style={line width=0.28pt},
        grid=both,
        minor tick num=1,
        major grid style={gray!50, dash pattern=on 0.8pt off 0.8pt, line width=0.16pt},
        minor grid style={gray!28, dash pattern=on 0.6pt off 0.6pt, line width=0.12pt},
        tick align=outside,
        tick pos=left,
        tick style={black, line width=0.12pt},
        yticklabel style={font=\fontsize{5.0}{4.4}\selectfont, xshift=1pt},
        xticklabel style={font=\fontsize{5.0}{4.4}\selectfont},
        ylabel style={font=\fontsize{5.4}{4.8}\selectfont, yshift=-0.05cm, xshift=-0.05cm},
        xlabel style={font=\fontsize{5.4}{4.8}\selectfont, yshift=0.03cm},
        label style={font=\fontsize{5.4}{4.8}\selectfont},
        title style={
            at={(0.5,-0.34)},
            anchor=north,
            font=\fontsize{5.6}{5.0}\selectfont\bfseries\itshape
        },
        every axis post/.append style={mark size=1.8pt, line width=0.7pt}
    ]

        % (a) Message dimension: MAE
        \nextgroupplot[
            ylabel={MAE},
            xmin=3.5, xmax=12.5,
            xtick={4,8,12},
            xticklabels={$4$,$8$,$12$},
            ymin=3.18, ymax=3.28,
            ytick={3.18,3.20,3.22,3.24,3.26,3.28},
            title={(a)}
        ]
        \addplot[color={rgb,255:red,86;green,180;blue,233}, mark=*] coordinates {
            (4,3.2442) (8,3.1921) (12,3.1838)
        };
        \addplot[color={rgb,255:red,230;green,159;blue,0}, mark=square*] coordinates {
            (4,3.2563) (8,3.2110) (12,3.1941)
        };

        % (b) Message dimension: RMSE
        \nextgroupplot[
            ylabel={RMSE},
            xmin=3.5, xmax=12.5,
            xtick={4,8,12},
            xticklabels={$4$,$8$,$12$},
            ymin=6.25, ymax=6.46,
            ytick={6.25,6.30,6.35,6.40,6.45},
            title={(b)}
        ]
        \addplot[color={rgb,255:red,86;green,180;blue,233}, mark=*] coordinates {
            (4,6.4044) (8,6.2781) (12,6.2941)
        };
        \addplot[color={rgb,255:red,230;green,159;blue,0}, mark=square*] coordinates {
            (4,6.4358) (8,6.2942) (12,6.3276)
        };

        % (c) Top-k sensitivity: MAE
        \nextgroupplot[
            ylabel={MAE},
            xmin=0.5, xmax=5.5,
            xtick={1,3,5},
            xticklabels={$1$,$3$,$5$},
            ymin=3.18, ymax=3.22,
            ytick={3.18,3.19,3.20,3.21,3.22},
            title={(c)}
        ]
        \addplot[color={rgb,255:red,0;green,158;blue,115}, mark=triangle*] coordinates {
            (1,3.1937) (3,3.1921) (5,3.2016)
        };

        % (d) Top-k sensitivity: RMSE
        \nextgroupplot[
            ylabel={RMSE},
            xmin=0.5, xmax=5.5,
            xtick={1,3,5},
            xticklabels={$1$,$3$,$5$},
            ymin=6.26, ymax=6.32,
            ytick={6.26,6.28,6.30,6.32},
            title={(d)}
        ]
        \addplot[color={rgb,255:red,0;green,158;blue,115}, mark=triangle*] coordinates {
            (1,6.3033) (3,6.2781) (5,6.2887)
        };

        % (e) Gradient-controller strength: MAE
        \nextgroupplot[
            ylabel={MAE},
            xmin=0.025, xmax=0.225,
            xtick={0.05,0.10,0.20},
            xticklabels={$0.05$,$0.10$,$0.20$},
            ymin=3.18, ymax=3.22,
            ytick={3.18,3.19,3.20,3.21,3.22},
            title={(e)}
        ]
        \addplot[color={rgb,255:red,204;green,121;blue,167}, mark=diamond*] coordinates {
            (0.05,3.2090)
            (0.10,3.1878)
            (0.20,3.1823)
        };

        % (f) Gradient-controller strength: RMSE
        \nextgroupplot[
            ylabel={RMSE},
            xmin=0.025, xmax=0.225,
            xtick={0.05,0.10,0.20},
            xticklabels={$0.05$,$0.10$,$0.20$},
            ymin=6.29, ymax=6.33,
            ytick={6.29,6.30,6.31,6.32,6.33},
            title={(f)}
        ]
        \addplot[color={rgb,255:red,204;green,121;blue,167}, mark=diamond*] coordinates {
            (0.05,6.3181)
            (0.10,6.3025)
            (0.20,6.2992)
        };
  \end{groupplot}
  \end{tikzpicture}

  \vspace{0.02cm}
  \caption{Hyperparameter ablation on METR-LA. Subfigures (a) and (b) vary the boundary-message dimension $d_m$ under two temporal decorrelation strengths. Subfigures (c) and (d) vary the number of received sender contexts top-$k$ using the default $\lambda_t=0.10$. Subfigures (e) and (f) vary the gradient-controller strength by jointly setting $\lambda_{\mathrm{align}}=\lambda_{\mathrm{div}}=\delta$.}
  \label{fig:hparam_ablation}
\end{figure}

Figure~\ref{fig:hparam_ablation} shows that FedeRICo is not highly sensitive to either boundary-message or gradient-controller hyperparameters. Increasing the message dimension from $d_m=4$ to $d_m=8$ improves both MAE and RMSE, while increasing to $d_m=12$ gives only marginal additional benefit. The two $\lambda_t$ curves remain close, indicating that performance is not tied to a narrowly tuned temporal decorrelation strength. The top-$k$ sweep is similarly stable, with one, three, or five sender contexts producing small changes. The gradient-controller sweep also remains within a narrow range, suggesting that branch-gradient coordination does not depend on a precisely tuned coupling strength.

\subsection{Client Scalability Comparison}

\begin{table}[t]
\centering
\normalsize
\caption{Federated model results on METR-LA across different numbers of clients.}
\label{tab:client_scaling_results}
\begin{tabular}{ccccc}
\toprule
\textbf{\# Clients} & \textbf{Metric} & \textbf{pFedCTP} & \textbf{FedDis} & \textbf{FedeRICo} \\
\midrule
\multirow{3}{*}{4}
& MAE  & 3.75 & 3.47 & 3.22 \\
& RMSE & 7.27 & 6.86 & 6.35 \\
& MAPE & 10.74 & 9.74 & 8.93 \\
\midrule
\multirow{3}{*}{8}
& MAE  & 3.80 & 3.49 & 3.16 \\
& RMSE & 7.27 & 6.91 & 6.24 \\
& MAPE & 10.85 & 9.88 & 8.76 \\
\midrule
\multirow{3}{*}{12}
& MAE  & 3.76 & 3.51 & 3.19 \\
& RMSE & 7.01 & 6.97 & 6.31 \\
& MAPE & 10.39 & 9.94 & 8.87 \\
\midrule
\multirow{3}{*}{16}
& MAE  & 3.79 & 3.55 & 3.26 \\
& RMSE & 7.21 & 7.04 & 6.45 \\
& MAPE & 11.02 & 10.08 & 9.11 \\
\midrule
\multirow{3}{*}{20}
& MAE  & 3.82 & 3.61 & 3.26 \\
& RMSE & 7.27 & 7.17 & 6.48 \\
& MAPE & 10.97 & 10.28 & 9.13 \\
\bottomrule
\end{tabular}
\end{table}

Table~\ref{tab:client_scaling_results} evaluates robustness to different numbers of METR-LA clients. FedeRICo consistently outperforms pFedCTP and FedDis across all partition sizes, showing that its boundary-message and gradient-coordination mechanisms remain effective as the graph is partitioned into more subsets for each client. This also demonstrates that FedeRICo scales better than the competing federated baselines under increasing client partitioning.

\subsection{Computational Costs Comparison}

\begin{table}[t]
\centering
\normalsize
\caption{Computational cost comparison on METR-LA with 12 clients. Parameter counts are reported as mean trainable parameters per client. Training time is the median logged wall-clock time per federated round. For FedeRICo, the two times denote sender-local encoder fitting and federated forecasting training, respectively.}
\label{tab:computation_costs}
\begin{tabular}{ccc}
\toprule
\textbf{Method} & \textbf{Param Count} & \textbf{Training Time/Round} \\
\midrule
MFVSTGNN  & 11.4K  & 40.6s \\
FedGTP    & 313.9K  & 1186.5s \\
pFedCTP   & 69.0K   & 58.6s \\
FedDis    & 1.52M  & 148.4s \\
FedeRICo    & 165.9K $+$ 1.3K/encoder & 21.3s / 40.7s \\
\bottomrule
\end{tabular}
\end{table}

Table~\ref{tab:computation_costs} compares the computational cost of the federated methods on METR-LA. FedeRICo uses more parameters than lightweight personalised baselines such as pFedCTP, but remains substantially smaller than FedDis and avoids the high per-round runtime of FedGTP. The additional sender-local encoder cost is separated from the main federated training loop, and the forecasting round time remains competitive with the other federated baselines. This indicates that the performance gains of FedeRICo do not come from simply scaling model size or incurring substantially higher federated training cost.

\subsection{Boundary Messaging Privacy Consideration}

We audit a compromised-receiver attack in which one client trains an inversion model using boundary-message examples for which it has corresponding local raw observations, and then transfers this model to incoming boundary messages from neighbouring clients. The attacker is implemented as a contextual MLP regressor, where the boundary packet is encoded by a two-layer MLP, temporal covariates are projected, receiver boundary-node identity and sender rank are embedded, and the fused representation is decoded to the raw boundary signal using an $\ell_1$ reconstruction loss.

\begin{table}[t]
\centering
\normalsize
\setlength{\tabcolsep}{3.5pt}
\caption{Compromised-receiver transfer reconstruction audit on METR-LA, $K=12$ Voronoi. Here, metadata denotes the temporal features, receiver boundary-node identity, and sender rank.}
\label{tab:boundary_privacy_audit}
\begin{tabular}{ccccc}
\toprule
\textbf{Input} & \textbf{MAE} & \textbf{RMSE} & \textbf{Corr.} & \textbf{$R^2$} \\
\midrule
Packet + metadata & 25.22 & 37.74 & 0.200 & -3.123 \\
Metadata only     & 15.80 & 23.57 & 0.084 & -0.190 \\
\bottomrule
\end{tabular}
\end{table}

The packet-conditioned attacker does not improve over the metadata-only baseline in this transfer setting. This suggests that sender-local boundary packets do not provide a stable inversion signal to a receiver trained only on its own locally observable examples.

\section{Conclusion}
This paper introduced FedeRICo, a federated spatial-temporal forecasting framework for traffic networks partitioned across heterogeneous client subgraphs. This approach separates collaboration from parameter sharing by coordinating clients through gradient-direction alignment while preserving client-specific forecasting parameters. It further restores cross-partition spatial context through boundary-aware residual messages exchanged between physically adjacent clients without sharing raw traffic observations.

Experiments on four real-world traffic benchmarks show that FedeRICo consistently outperforms existing federated spatial-temporal baselines and narrows the gap to centralised forecasting. Ablation, scalability, hyperparameter, and reconstruction-audit results further indicate that boundary residual messaging and gradient-level coordination are complementary, robust across client partitions, and do not provide a stable inversion signal under the compromised-receiver attack considered.

Future work may aim extend the framework for asynchronous participation and stronger formal privacy mechanisms for boundary-message exchange.

\phantomsection
\section*{GenAI Disclosure Statement}
Use of Generative AI for this work was solely for language refinement, formatting assistance, and grammatical corrections. 

\bibliographystyle{ACM-Reference-Format}
\bibliography{sample-base}

\end{document}